\documentclass[sigconf]{acmart}
\usepackage{balance}
\usepackage{enumitem}
\AtBeginDocument{%
  \providecommand\BibTeX{{%
    \normalfont B\kern-0.5em{\scshape i\kern-0.25em b}\kern-0.8em\TeX}}}

\copyrightyear{2021}
\acmYear{2021}
\setcopyright{acmlicensed}\acmConference[ICCPS '21]{ACM/IEEE 12th International Conference on Cyber-Physical Systems (with CPS-IoT Week 2021)}{May 19--21, 2021}{Nashville, TN, USA}
\acmBooktitle{ACM/IEEE 12th International Conference on Cyber-Physical Systems (with CPS-IoT Week 2021) (ICCPS '21), May 19--21, 2021, Nashville, TN, USA}
\acmPrice{15.00}
\acmDOI{10.1145/3450267.3450543}
\acmISBN{978-1-4503-8353-0/21/05}

\usepackage{amsmath}
\usepackage{appendix}
\usepackage{amsfonts}
\usepackage{graphicx}
\usepackage{xcolor}
\usepackage{bbm}
\usepackage[ruled,vlined,linesnumbered]{algorithm2e}
\SetArgSty{textnormal}
\usepackage[normalem]{ulem}
\usepackage[ruled,vlined]{algorithm2e}
\usepackage{caption}
\usepackage{subcaption}
\usepackage{soul}

\usepackage{pgfplots}
\pgfplotsset{compat=1.8}

\usepgfplotslibrary{statistics}
\makeatletter
\pgfplotsset{
    boxplot/hide outliers/.code={
        \def\pgfplotsplothandlerboxplot@outlier{}%
    }
}
\makeatother

\usepackage{color, colortbl}
\usepackage{bbold}

\DeclareMathOperator*{\argmax}{arg\,max}

\usepackage[textsize=tiny, textwidth=1.23cm]{todonotes}

\begin{document}
\settopmatter{printacmref=false}
\setcopyright{none}
\renewcommand\footnotetextcopyrightpermission[1]{}
% \section*{Misc}
% need for cross referencing to work in overleaf.will remove later.
% \newpage
% \pagenumbering{arabic}

%%
%% The "title" command has an optional parameter,
%% allowing the author to define a "short title" to be used in page headers.
\title{Hierarchical Planning for Resource Allocation \\ in Emergency Response Systems}

\author{Geoffrey Pettet}
\affiliation{%
  \institution{Vanderbilt University}
  \city{Nashville}
  \state{TN}
  \postcode{37212}
}
\email{geoffrey.a.pettet@vanderbilt.edu}

\author{Ayan Mukhopadhyay}
\affiliation{%
  \institution{Vanderbilt University*}
  \city{Nashville}
  \state{TN}
  \postcode{37212}
}
\email{ayan.mukhopadhyay@vanderbilt.edu}

\author{Mykel J. Kochenderfer}
\affiliation{%
  \institution{Stanford University}
  \city{Stanford}
  \state{CA}
  \postcode{94305}
}
\email{mykel@stanford.edu}

\author{Abhishek Dubey}
\affiliation{%
  \institution{Vanderbilt University}
  \city{Nashville}
  \state{TN}
  \postcode{37212}
}
\email{abhishek.dubey@vanderbilt.edu}

\thanks{This work is sponsored by The National
Science Foundation under award numbers CNS1640624 and IIS1814958 and a grant from Tennessee Department of Transportation. We also acknowledge support from Google through the Cloud Research Credits. \\
* The author was at Stanford University when this work started and was sponsored by the Center of Automotive Research at Stanford.}

\begin{abstract}
    A classical problem in city-scale cyber-physical systems (CPS) is resource allocation under uncertainty. Typically, such problems are modeled as Markov (or semi-Markov) decision processes. While online, offline, and decentralized approaches have been applied to such problems, they have difficulty scaling to large decision problems. We present a general approach to hierarchical planning that leverages structure in city-level CPS problems for resource allocation under uncertainty. We use emergency response as a case study and show how a large resource allocation problem can be split into smaller problems. We then create a principled framework for solving the smaller problems and tackling the interaction between them. Finally, we use real-world data from Nashville, Tennessee, a major metropolitan area in the United States, to validate our approach. Our experiments show that the proposed approach outperforms state-of-the-art approaches used in the field of emergency response. 
\end{abstract}

%%
%% The code below is generated by the tool at http://dl.acm.org/ccs.cfm.
%% Please copy and paste the code instead of the example below.
%%
% \begin{CCSXML}
% <ccs2012>
% <concept>
% <concept_id>10002951.10003227.10003241</concept_id>
% <concept_desc>Information systems~Decision support systems</concept_desc>
% <concept_significance>500</concept_significance>
% </concept>
% <concept>
% <concept_id>10002951.10003227.10003236</concept_id>
% <concept_desc>Information systems~Spatial-temporal systems</concept_desc>
% <concept_significance>300</concept_significance>
% </concept>
% <concept>
% <concept_id>10002950.10003648.10003700.10003701</concept_id>
% <concept_desc>Mathematics of computing~Markov processes</concept_desc>
% <concept_significance>300</concept_significance>
% </concept>
% <concept>
% <concept_id>10010405.10010481.10010484</concept_id>
% <concept_desc>Applied computing~Decision analysis</concept_desc>
% <concept_significance>500</concept_significance>
% </concept>
% </ccs2012>
% \end{CCSXML}

% \ccsdesc[500]{Information systems~Decision support systems}
% \ccsdesc[300]{Information systems~Spatial-temporal systems}
% \ccsdesc[300]{Mathematics of computing~Markov processes}
% \ccsdesc[500]{Applied computing~Decision analysis}

%%
%% Keywords. The author(s) should pick words that accurately describe
%% the work being presented. Separate the keywords with commas.

%% A "teaser" image appears between the author and affiliation
%% information and the body of the document, and typically spans the
%% page.

%%
%% This command processes the author and affiliation and title
%% information and builds the first part of the formatted document.
\maketitle
 \begin{center}
 \textbf{Accepted for publication in the proceedings of the 12th ACM/IEEE International Conference on Cyber-Physical Systems (ICCPS-2021)}.
 \end{center}\vspace{1em}
\vspace{-0.2in}
\section{Introduction}

Dynamic resource allocation (DRA) in anticipation of uncertain demand is a canonical problem in city-scale cyber-physical systems (CPS)~\cite{MukhopadhyayICCPS}. In such a scenario, the decision-maker optimizes the spatial location of resources (typically called \textit{agents}) to maximize utility over time while satisfying constraints specific to the domain of the CPS. Consider the problem of emergency response. Governments and private agencies optimize ambulance locations to minimize response times to emergency calls while considering constraints on the number of available ambulances and locations where they can be stationed. An associated problem is resource dispatch, in which agents must spatially move from their location to the location of the demand to address the task at hand. Ambulances must move to the scene of the incidents to provide assistance to patients. This paper addresses the problem of dynamic resource allocation and dispatch in large-scale systems. 

DRA manifests in many problems at the intersection of urban management, CPS, and multi-agent systems. These include positioning electric scooters~\cite{gossling2020integrating}, optimizing electric vehicle charging station locations~\cite{baouche2014efficient}, designing on-demand transit~\cite{chebbi2015modeling}, and emergency response management (ERM)~\cite{mukhopadhyay2020review}. We focus on ERM for several reasons. First, it is a critical problem faced by communities across the globe. Responders must attend to many incidents dispersed across space and time using limited resources. \citeauthor{mukhopadhyay2020review}~\cite{mukhopadhyay2020review} describe the intricate pipeline that first-responders follow to ensure timely and effective service. Second, ERM pipelines are an important example of human-in-the-loop CPS (H-CPS), which introduces structure and constraints. Finally, emergency response presents us the scope to evaluate multi-agent resource allocation and dispatch problems with high-quality real-world data.

\begin{figure}[t]
    \centering
    \includegraphics[width=0.9\columnwidth]{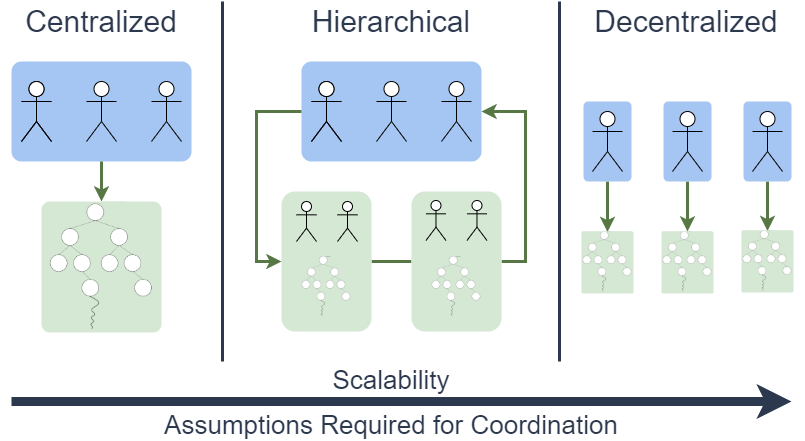}
    \caption{Spectrum of approaches to solve a dynamic resource allocation problem under uncertainty. A completely centralized approach deals with a monolithic state representation. In a completely decentralized approach, each agent simulates what other agents do and performs its own action estimation. Our hierarchical approach segments the planning problem into sub-problems to improve scalability without agents estimating other agents' actions.}
    \label{fig:planning}
    \vspace{-0.2in}
\end{figure}

DRA and dispatch problems are typically modeled as Markov decision processes (MDP). The decision-maker's goal is to find an optimal \textit{policy}, which is a mapping between system states and actions to be taken. 
% Problems pertaining to CPS in urban areas evolve in continuous time. The integration of dispatch into the problem makes state-transitions non-memoryless, so the dynamics of the underlying continuous-time stochastic process are actually semi-Markovian~\cite{mukhopadhyayAAMAS18}. 
There is a broad spectrum of approaches available to address resource allocation under uncertainty, as shown in figure ~\ref{fig:planning}. In previous work, we applied the extremes of this continuum to ERM, with each having strengths and weaknesses~\cite{mukhopadhyayAAMAS18, MukhopadhyayICCPS, pettet2020algorithmic}. 

% \begin{figure*}[t]
%     \centering
%     \includegraphics[width=0.8\textwidth]{figures/PlanningParadigm.png}
%     \caption{Spectrum of approaches to solve a dynamic resource allocation problem under uncertainty. A completely centralized approach (leftmost) deals with a monolithic state representation. In a completely decentralized approach each agent simulates what other agents do and performs its own action estimation (rightmost). Our approach (middle) segments the original planning problem into multiple sub-problems to improve scalability without agents estimating other agents' actions.}
%     \label{fig:planning}
% \end{figure*}

%%% Old caption from figure above
% Different approaches to solve a dynamic resource allocation problem under uncertainty. A completely centralized approach (leftmost) deals with a monolithic state representation. Using such a representation, the decision-maker could estimate state transition probabilites using a simulator~\cite{mukhopadhyayAAMAS18} or use an online solution approach~\cite{pettet2020algorithmic}. A completely decentralized approach involves each agent simulating what other agents do and performing its own action estimation (rightmost). Our approach (shown in the middle) segments the original planning problem into multiple sub-problems to improve scalability without agents estimating other agents' actions.

The most direct approach is to represent the problem as a single MDP, shown on the left in figure ~\ref{fig:planning}. When real-world problems are modeled as MDPs, state transitions are unknown. The standard method to address this is to use a simulator to estimate an empirical distribution over the state transitions~\cite{mukhopadhyayAAMAS18}, and use this to learn a policy using the well-known policy iteration algorithm~\cite{kochenderfer2015decision}. Unfortunately, this offline approach does not scale to realistic problems~\cite{MukhopadhyayICCPS}. Another method is to use an online solution like Monte-Carlo tree search (MCTS)~\cite{MukhopadhyayICCPS}. While adaptable to dynamic environments, this still suffers from poor scalability, taking too long to converge for realistic scenarios.  

% The standard approach to address this issue is to use a black-box simulator which is relatively simple to construct~\cite{mukhopadhyayAAMAS18}. Then, the simulator can be used to estimate an empirical distribution over the state transitions~\cite{mukhopadhyayAAMAS18}. Given the transition distribution, an optimal policy can be learned using the well-known policy iteration algorithm~\cite{kochenderfer2015decision}. Another approach is to use an online solution, where given a specific state, the simulator aids a heuristic search algorithm like Monte-Carlo tree search (MCTS)~\cite{MukhopadhyayICCPS}.

On the other side of the spectrum is a completely decentralized methodology, as shown in the extreme right in figure ~\ref{fig:planning}. In such an approach, each agent determines its own course of action. As the agents cooperate to achieve a single goal, they must estimate what other agents will do in the future as they optimize their own actions. For example, \citeauthor{claes2017decentralised}~\citep{claes2017decentralised} show how each agent can explore the efficacy of its actions locally by using MCTS. While such approaches are significantly more scalable than their centralized counterparts, they are sub-optimal as agents' estimates of other agents' actions can be highly inaccurate. Note that high-fidelity models for estimating agents' actions limits scalability and therefore decentralized approaches rely on inexpensive heuristics. Decentralized approaches are useful in disaster scenarios where communication networks can break down, 
% In such cases, agents must plan for themselves with locally available information.  
but agents in urban areas (ambulances) typically have access to reliable networks and communication is not a constraint. Therefore, approaches that ensure scalability but do not fully use available information during planning are not suitable for emergency response in urban areas, especially when fast and effective response is critical.  

We explore hierarchical planning~\cite{hauskrecht2013hierarchical}, which focuses on learning local policies, known as \textit{macros}, over subsets of the state space. We use hierarchical planning to address resource allocation for emergency response by leveraging spatial structure in the problem. Our idea is also motivated by the concept of \textit{jurisdictions} or \textit{action-areas} used in public policy, which create different zones to partition and better manage infrastructure. 

We design a principled algorithmic approach that partitions the spatial area under consideration. We then treat resource allocation in each resulting sub-area (called regions) as individual planning problems which are smaller than the original problem by construction. While this ensures scalability, it naturally results in performance loss because agents constrained in one region might be needed in the other region. We show how hierarchical planning can be used to facilitate transfer of agents across regions. A top-level planner, called the inter-region planner, identifies and detects states where ``interaction'' between regions is necessary and finds actions such that the overall utility of the system can be maximized. A low-level planner, called the intra-region planner, addresses allocation and dispatch within a region.

\textbf{Contributions}: \textbf{1)} We leverage structure in resource allocation problems to design a hierarchical planning approach that scales significantly better than prior approaches. The key idea in our approach comes from the concept of \textit{macros} in decision-theoretic systems~\cite{hauskrecht2013hierarchical}, which focus on finding policies for subsets of the state-space, thereby ensuring scalability. \textbf{2)} We show how exogenous constraints in real-world resource allocation problems can be used to naturally partition the overall decision-problem into sub-problems. We create a low-level planner that focuses on finding optimal policies for the sub-problems. \textbf{3)} We show how a high-level planner can facilitate exchange of resources between the sub-problems (spatial areas in our case). \textbf{4)} We use real-world emergency response data from Nashville, Tennessee, a major metropolitan area in the United States, to evaluate our approach and show that it performs better than state-of-the-art approaches both in terms of efficiency, scalability, and robustness to agent failures.

The paper is organized as follows. We describe ERM and a mathematical formulation of our problem in section~\ref{sec:problem_formulation}. Section~\ref{sec:approach} describes the overall approach, the high-level, and the low-level planner. We present our experimental design in section~\ref{sec:exp} and the results in section~\ref{sec:exp_results}. We present related work in section~\ref{sec:related} and summarize the paper in section~\ref{sec:conclusion}. Additionally, an appendix discusses reproducibility, notation, and implementation. 

%we are not really giving details :)

\section{Problem Formulation}\label{sec:problem_formulation}

% We deal with the problem of spatial-temporal resource allocation in anticipation of demand for a specific service. For example, consider the problem of emergency response in urban areas (ERM). A set of agents (ambulances in this case) must be positioned such that they can quickly respond to medical calls across a geographical area. 

%\hl{DO AT THE END --- figures need to be rearranged so that they appear close to where they are being referred to.}
Emergency response management (ERM) involves responding to spatial-temporal calls for service in a specified spatial area. The agents, ambulances in this case, respond to calls for medical aid. Once an incident  is reported, responders  are  dispatched  by  a human agent to the scene of the incident (guided by some algorithmic approach). 
% This process typically takes a few seconds in practice, but can take longer if all responders are busy. 
If no free responder is available, the incident typically enters a waiting queue, and is responded to when an agent becomes free (we use ``agent'' and ``responder'' inter-changeably throughout the paper). Each responder is typically housed at specific locations called depots, which are distributed in the spatial area under consideration (these could be fire stations or rented parking spots, for example). Once a responder finishes servicing an incident, it is directed back to a depot and becomes available for dispatch. Therefore, there are two broad actions that the decision maker can optimize: (1) which responder to dispatch once an incident occurs (dispatching action) and (2) which depots to send the responders to in anticipation of future incidents (allocation action).

To model the problem of emergency response management, we begin with several assumptions on the problem structure and information provided \textit{a priori}. First, we assume that we are given a spatial map broken up into a finite collection of equally sized cells $G$, a set of agents $\Lambda$ that need to be allocated across these cells and dispatched to demand points, and a travel model that describes how the agents move throughout $G$. We also assume that we have access to a spatial-temporal model of demand over $G$, and that within each cell the temporal demand distribution is homogeneous. Our third assumption is that agent allocation is restricted to \textit{depots} $D$, that are located in a fixed subset of cells. Each depot $d \in D$ has a fixed capacity $\mathcal{C}(d)$, which is the number of agents it can accommodate.

While the state space in this resource allocation problem evolves in continuous-time, it is convenient to view the dynamics as a set of finite decision making states that evolve in discrete time. For example, an ambulance moving through an area continuously changes the state of the \textit{world}, but presents no scope for decision-making unless an event occurs that needs response or the planner redistributes agents. As a result, the decision-maker only needs to find optimal actions for a subset of the state space that provides the opportunity to take actions. 

A key component of response problems is that agents physically move to the site of the request, which makes temporal transitions between decision-making states non-memoryless. This causes the underlying stochastic process governing the evolution of the system to be semi-Markovian. The dynamics of a set of agents working to achieve a common goal can be modeled as a Multi-Agent Semi-Markov Decision Process (MSMDP)~\cite{rohanimanesh2003learning}, which can be represented as the tuple $(S,\Lambda,\mathcal{A},P,T,\rho(s,a),\alpha,\mathcal{T})$, where $S$ is a finite state space, $\rho(s, a)$ represents the instantaneous reward for taking action $a$ in state $s$, $P$ is a state transition function, $T$ is the temporal distribution over transitions between states, $\alpha$ is a discount factor, and $\Lambda$ is a finite collection of agents where $\lambda_j \in \Lambda$ denotes the $j$th agent. The action space of the $j$th agent is represented by $A_j$, and $\mathcal{A} = \prod_{i=1}^{m} A_j$ represents the joint action space of all agents. We assume that the agents are cooperative and work to maximize the overall utility of the system. $\mathcal{T}$ represents a termination scheme; note that since agents each take different actions that could take different times to complete, they may not all terminate at the same time~\cite{rohanimanesh2003learning}. We focus on asynchronous termination, where actions for a particular agent are chosen as and when the agent completes its last assigned action.\footnote{Different termination schemes are discussed in the theoretical analysis by ~\citeauthor{rohanimanesh2003learning}~\cite{rohanimanesh2003learning}.}

\begin{figure}[t]
    \centering
    \includegraphics[width=0.75\columnwidth]{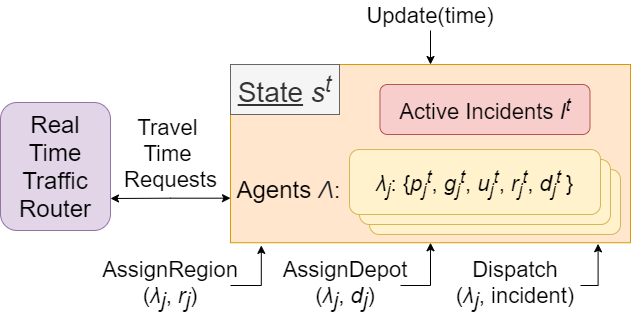}
    \caption{System State and Actions.}
    \label{fig:state}
    \vspace{-0.2in}
\end{figure}

\textbf{States:} A state at time $t$ is represented by $s^t$ and consists of a tuple $(I^t, \mathcal{Q}(s^t))$, where $I^t$ is a collection of cell indices that are waiting to be serviced, ordered according to the relative times of incident occurrence. $\mathcal{Q}(s^t)$ corresponds to information about the set of agents at time $t$ with $|\mathcal{Q}(s^t)| = |\Lambda_r|$. Each entry $q^t_j \in \mathcal{Q}(s^t)$ is a set $\{p^t_j,g^t_j,u^t_j, r_j^t, d_j^t\}$, where $p^t_j$ is the position of agent $\lambda_j$, $g^t_j$ is the destination cell that it is traveling to (which can be its current position), $u^t_j$ is used to encode its current status (busy or available), $r_j^t$ is the agent's assigned region, and $d_j^t$ is its assigned depot, all observed at time $t$. A diagram of the state is shown in figure \ref{fig:state} and discussed in detail in appendix \ref{sec:app_sim}.

We assume that no two events occur simultaneously in our system model. In such a case, since the system model evolves in continuous time, we can add an arbitrarily small time interval to create two separate states. 

\textbf{Actions:}
Actions correspond to directing agents to valid cells to either respond to demand or wait at a depot. For a specific agent $\lambda_i \in \Lambda_r$, valid actions for a specific state $s_i$ are denoted by $A^i(s_i)$ (some actions are naturally invalid, for example, if an agent is at cell $k$ in the current state, any action not originating from cell $k$ is unavailable to the agent). Actions can be broadly divided into two categories: \textit{dispatching} actions which direct agents to service an active demand point and \textit{allocation} actions which assign agents to wait in particular depots in anticipation of future demand. 
%\hl{AD: can you introduce definition environment to define dispatching and allocation and reference it later. Ayan: I am not quite sure what this comment means.}

Here it is important to note that a key aspect of emergency response is that if any free responders are available when an incident is reported, then the nearest one must be greedily dispatched to attend to the incident. This constraint is a direct consequence of the bounds within which emergency responders operate, as well as the critical nature of the incidents~\cite{mukhopadhyay2020review, mukhopadhyay2020designing, pettet2020algorithmic}. Therefore, the problem we consider focuses on proactively redistributing agents across a spatial area under future demand uncertainty. Nonetheless, dispatch actions are still necessary to model since they are the foundation of our reward function.

% Here it is important to note that in the ERM case study, dispatch actions are greedily prescribed~\cite{mukhopadhyay2020review,pettet2020algorithmic}. In practice, the severity of incidents can not be gauged from calls for service and therefore, the closest available responder must be dispatched to the incidents. We consider that dispatch actions are greedy by default, and therefore do not optimize over such actions.  
% The joint valid action space of all the agents in a region $r$ and a particular instantiation of it are defined by $\mathcal{A}_r$ and $a$ respectively, and that of a specific agent $\lambda_j$ by $A_j$ and $a_j$.

\textbf{Transitions:}
The system model of ERM evolves through several stochastic processes. Incidents occur at different points in time and space governed by some arrival distribution. We assume that the number of incidents in a cell $r_j \in R$ per unit time can be approximated by a Poisson distribution with mean rate $\gamma_j$ (per unit time), a commonly used model for spatial-temporal incident occurrence~\cite{mukhopadhyay2020review}. Agents travel from their locations to the scene of incidents governed by a model of travel times. We assume that agents then take time to service the incident at some exogenously specified velocity. The system itself takes time to plan and implement allocation of responders. We refrain from discussing the mathematical model and expressions for the temporal transitions and the state transition probabilities, as our algorithmic framework only needs a generative model of the world (in the form of a black-box simulator) and not explicit estimates of transitions themselves.

% Having described the evolution of our world, we now look at both the transition time between states, as well as the probability of observing a state, given the last state and action taken. We define the former first, denoting the time between two states $s_i$ and $s_j$ by the random variable $t_{ij}$. There are four random variables of interest in this context. We denote the time between incidents by the random variable $t_a$, the time to service an incident by $t_s$, the time taken for an allocation step as $t_b$ and the time taken for a responder to reach the scene of an incident by $t_r$.  
% Specifically, we model $t_a$ using a survival model described in section \ref{sec:survival}. We model the service times ($t_s$) by learning an exponential distribution from service times using historical emergency response data, and we model allocation time ($t_b$) simply by the time taken by an agent to move to the directed depot.

\textbf{Rewards:}
Rewards in an SMDP usually have two components: a lump sum instantaneous reward for taking actions, and a continuous time reward as the process evolves. Our system only involves the former, which we denote by $\rho(s, a)$, for taking action $a$ in state $s$. Rewards are highly domain dependent; the metric we are concerned with in our ERM case study is \textit{incident response time} $t_r$, which is the time between the system becoming aware of an incident and when the first agent arrives on scene. 
% Unlike previous work in this area that was also concerned with minimizing the distance traveled by responders for allocation \cite{pettet2020algorithmic}, we are not concerned with this since responders are limited to moving within their small region. 
Therefore, our reward function is
\begin{equation}\label{eq:running action_util}
        \rho(s,a) = 
        \begin{cases}
        \alpha^{t_h}(t_{r}(s,a)) & \text{if dispatching} \\
        0                             & \text{otherwise}
        \end{cases}
\end{equation}
where $\alpha$ is the discount factor for future rewards, $t_h$ the time since the beginning of the planning horizon $t_{0}$, and $t_{r}(s,a)$ is the response time to the incident due to a dispatch action. The benefits of allocation actions are inferred from improved future dispatching.

\textbf{Problem Definition}
 Given state $s$ and a set of agents $\Lambda$, our goal is to find an action recommendation set $\sigma = \{a_1,...,a_m\}$ with $a_i \in A^i(s)$ that maximizes the expected reward. The $i$th entry in $\sigma$ contains a \textit{valid} action for the $i$th agent. In our ERM case study, this corresponds to finding an allocation of agents to depots that minimizes the expected response times to incidents.
 
 \begin{figure}[t]
    \centering
    \includegraphics[width=1.0\columnwidth]{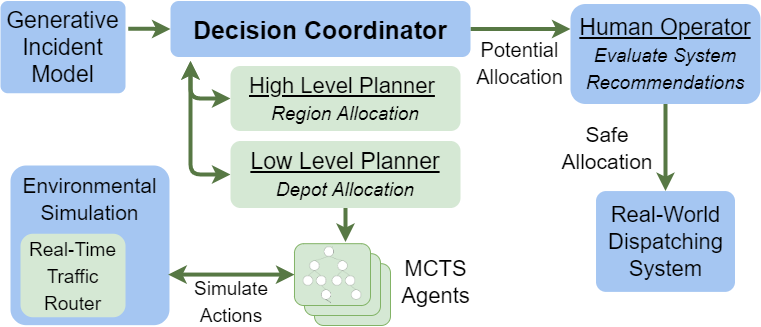}
    \caption{Solution Approach}
    \label{fig:software_diagram}
    \vspace{-0.2in}
\end{figure}

\section{Approach}\label{sec:approach}

We show a schematic representation of our decision support system in Fig. \ref{fig:software_diagram}. We assume the availability of historical incident data and a probabilistic generative model of incident occurrence learned from the data. We also assume access to a traffic router that can simulate traffic in the concerned urban area (for example, see prior work done by ~\citeauthor{MukhopadhyayICCPS}~\cite{MukhopadhyayICCPS}). We divide tasks pertaining to planning in a hierarchical manner. We refer to the two stages in our hierarchy as the ``high-level'' and the ``low-level''. The high-level planner divides the overall decision-theoretic problem into smaller sub-problems by creating meaningful spatial clusters, which we call regions. It also optimizes the distribution of agents among the regions. Then, an instance of the low-level planner is instantiated for each of the regions. The low-level planner for a specific region optimizes the spatial locations of agents within that region. Finally, a human operator accesses the planning mechanism to act as an interface with a real-world computer-aided dispatch system. 

Segmenting allocation into smaller sub-problems significantly reduces its complexity compared to a centralized problem structure. Consider an example city with $|\Lambda|=20$ responders and $|D|=30$ depots, each of which can hold one responder. With a centralized approach, any responder can go to any depot, so there are $\text{\textit{Permutations}}(|D|,|\Lambda|) = \frac{|D|!}{(|D|-|\Lambda|)!} = \frac{30!}{10!} = 7.31\times10^{25}$ possible assignments at each decision epoch. Now consider a hierarchical structure where the problem is split into $5$ evenly sized sub-problems, so $|\Lambda_h|=4$ and $|D_h| = 6$ for each region. There are now $P(|D_h|,|\Lambda_h|) = 360$ possible allocations in each region, so there are $360*5=1800$ possible actions across all regions. This is a reduction in complexity of about $22$ orders of magnitude compared to the centralized problem. Alternatively, a decentralized approach in which each responder plans only its own actions reduces the complexity further to only $|D|=30$ possible allocations for each responder~\cite{pettet2020algorithmic}. However, this reduction in complexity comes at a cost, as each agent must make assumptions regarding other responder behavior, which can lead to sub-optimal planning. Hierarchical planning offers a balance between decentralized and centralized planning by having agents accurately model other nearby agents' behavior while having reasonable decision complexity.

% experiments. It has $|D|=35$ depots and $|\Lambda|=26$ responders. Centralized: Any responder can go to any depot, so the number of possible allocations are $P(|D|,|\Lambda|) = \frac{|D|!}{(|D|-|\Lambda|)!} = \frac{35!}{9!} = 2.85e+34$. Decentralized: each responder builds it's own decision tree focused only on it's allocation, so there are $|D|$ possible actions. Hierarchical: Here, each region has it's own MDP, so each MDP's complexity depends on the region definitions. In our expierements below, our smallest region had $|D_r|=2$ assigned depots and $|\Lambda_r|=1$ assigned responders, giving $\num{2}$
% possible allocations. Our largest region had $|D_r|=11$ assigned depots and $|\Lambda_r|=8$ assigned responders, giving $6.65e+6$ possible allocations. This means that our worst case region decreased the action complexity by $\sim 27$ orders of magnitude.

An important consideration in designing approaches to resource allocation under uncertainty in city-level CPS problems is to adapt to the dynamic environment in which such systems evolve. In our decision support system, a decision coordinator (an automated module) invokes the high-level planner at all states that allow the scope for making decisions. For example, consider that a responder is unavailable due to maintenance. The coordinator triggers the high-level planner and notifies it of the change. The high-level planner then checks if the spatial distribution of the responders can be optimized to best respond to the situation at hand. We describe the exact optimization functions, metrics, and approaches that we use to design the planners below.

\subsection{High-Level Planner}

Through the high-level planner, we seek to decompose the overall MSMDP into a set of smaller problems that can be solved tractably and distribute agents among the regions. Recall that the overall goal of the parent MSMDP is to reduce expected response times. Response times to emergency incidents consist of two parts: a) the time taken by an agent to travel to the scene of the incident, and b) the time taken to service the incident. We assume that incidents are homogeneous, meaning that the time taken by responders to service incidents follows the same distribution. This assumption means that the sole criterion that a planner needs to optimize is overall travel time of the agents to incidents 
% so that calls for service do not have to wait for agents to arrive on scene 
(achieving zero waiting times is clearly infeasible in practice, so we seek to minimize waiting times). Therefore, the high-level planner seeks to distribute responders to different regions such that incident waiting time is minimized. There are other factors that contribute to waiting times (traffic, for example). However, such factors are not affected by the distribution of the agents themselves.

Consider that the high-level planner seeks to divide the overall problem in to $k$ regions, denoted by the set $R = \{r_1,r_2,\dots , r_k\}$, where $r_i \in R$ denotes the $i$th region. To achieve this, we use historical data of incident occurrence to partition the set $G$ into clusters using a standard clustering algorithm (we use the $k$-means algorithm in our experiments). We denote the average waiting time for incidents in region $r_j \in R$ by $w_{j}(x_j)$, where $x_j$ is the number of responders assigned to the region $r_j$. We model waiting times in a region by a multi-server queue model. Recall that incident arrivals are distributed according to a Poisson distribution, thereby making inter-incident times exponentially distributed. We make the standard assumption that service times are exponentially distributed~\cite{mukhopadhyayAAMAS17}. One potential issue with using well-known queuing models to estimate waiting times in emergency response is that travel times are not memoryless. We use an approximation from prior work to tackle this problem~\cite{mukhopadhyayAAMAS17}. Specifically, travel times to emergency incidents are typically much smaller than service times. Thus, the sum of travel times and service times can be considered to be approximated by a memoryless distribution (provided that the service time itself can be modeled by a memoryless distribution). The average waiting-time for a region $r_j \in R$ can then be estimated by considering a $m/m/c$ queuing model (using Kendall's notation~\cite{kendall1953stochastic}), where $c=x_j$. 

Given estimated waiting times for incidents in each region, the high-level planner seeks to minimize the cumulative response times across all regions. The optimization problem can be represented as
\begin{subequations}
\label{eq:highLevel}
\begin{align}
	&\min_{x}  \sum_{j=1}^{k} w_j(x_j)\\
	&\text{s.t.}\,\,\,\,\sum_{i=1}^{k} x_i = |\Lambda| \label{cons:budget}\\
	& x_i \in \mathbb{Z}^{0+} \,\, \forall i \in \{1,\dots,k\}
\end{align}	
\end{subequations}

Let the average rate of incident occurrence be $\lambda_j$ at region $r_j \in R$. Let the average service time be $T_s$ and let $\mu=1/T_s$ denote the mean service rate. Then, the mean waiting time $w_j(x_j)$ is ~\cite{shortle2018fundamentals}: 
\[
    w_j(x_j) = \frac{P_0 (\frac{\lambda}{\mu})^{x_j}\lambda}{c!(1-\rho)^2 c}
\]
where $P_0$ denotes the probability that there are 0 incidents waiting for service and can be represented as
\[
    P_0 = 1/\Big[\sum_{m=0}^{x_j -1} \frac{(x_j \rho)^m}{m!} + \frac{(c \rho)^c}{c!(1 - \rho)} \Big]
\]

The objective function in mathematical program \ref{eq:highLevel} is non-linear and non-convex. We use an iterative greedy approach shown in algorithm \ref{algo:high}. We begin by sorting regions according to total arrival rates. Since the arrival process is assumed to be Poisson distributed, the overall rate for region $r_j \in R$, denoted by $\gamma_j$ can be calculated as $\sum_{g_i \in G} \mathbb{1}(g_i \in r_j)\gamma_i$, where $\mathbb{1}(g_i \in r_j)$ denotes an indicator function that checks if cell $g_i$ belongs to region $r_j \in R$. Let this sorted list be $R_s$. Then, we assign responders iteratively to regions in order of decreasing arrival rates (step \ref{algo:high_assign}). After assigning each responder to a region $r_j \in R$, we compare the overall service rate ($x_j$ times the mean service rate by one responder) and the incident arrival rate for the region (step \ref{algo:high_sustain}). Essentially, we try to ensure that given a pre-specified service rate, the expected length of the queue is not arbitrarily large. Once a region is assigned enough responders to sustain the arrival of incidents, we move on the next region in the sorted list $R_s$ (step \ref{algo:high_next}). Once all regions are assigned responders in this manner, we check if there are surplus responders (step \ref{algo:high_surplus}). The surplus responders are assigned iteratively according to the incremental benefit of each assignment. Specifically, for each region, we calculate the marginal benefit $J$ of adding one agent to the existing allocation (step \ref{algo:high_marginal}). Then, we assign an agent to the region that gains the most (in terms of reduction in waiting times) by the assignment.

\setlength{\textfloatsep}{5pt}
\begin{algorithm}[t]
\SetAlgoLined
\SetKwInOut{Input}{input}
\SetKwInOut{Output}{output}
\Input{Sorted Regions $R_s$, Arrival Rates $\{\gamma_1,\gamma_2,\dots,\gamma_k\}$, Service Rate $\eta$}
\Output{Responder Allocation $X\,=\,\{x_1,x_2,\dots,x_k\}$}
 $\text{assigned} := 0, i := 0, J := \emptyset$\;
 \While{\text{assigned} $\leq \,\,\mid \Lambda \mid \,\, \textbf{and} \,\, i \leq k$}{
  $x_i := x_i + 1$\;\label{algo:high_assign}
  $\text{assigned} := \text{assigned} + 1$\;
  \If{$\eta \times (x_i) \geq \sum_{g_i \in G} \mathbb{1}(g_i \in r)\gamma_i$ }{\label{algo:high_sustain}
  $i := i + 1$\;\label{algo:high_next}
  }
 }
 \While{\text{assigned} $\leq \,\,\mid \Lambda \mid$}{\label{algo:high_surplus}
  \For{$i \in [1,k]$}{$J[i] := w_i(x_i) - w_i(x_i + 1)$\;}
%   $\text{Calculate } J \text{ where } J_i = w_i(x_i) - w_i(x_i + 1)$\;
  $r^{*} := \argmax_{i \in [1,k]}J[i]$\;\label{algo:high_marginal}
  $x_{r^{*}} := x_{r^{*}} + 1$\;
  $\text{assigned} := \text{assigned} + 1$\;
 }
 \caption{High-Level Planner}
 \label{algo:high}
 %\vspace{-0.2in}
\end{algorithm}

% notes: 
%  also called the intra-region planner

\subsection{Low-Level Planner}

% The high-level planner's responsibility is to distribute agents across the regions, but 
The fine-grained allocation of agents to depots within each region is managed by the low-level planner, which induces a decision process for each region that is smaller than the original MSMDP problem described in section \ref{sec:problem_formulation} by design. The MSMDP induced by each region $r_j \in R$ contains only state information and actions that are relevant to $r_j$, i.e. the depots within $r_j$, the agent's assigned to $r_j$ by the inter-region planner, and incident demand generated within $r_j$. 

Decomposing the overall problem makes each region's MSMDP tractable using many approaches, such as dynamic programming, reinforcement learning (RL), and Monte Carlo Tree Search (MCTS). Each approach has advantages and tradeoffs which must be examined to determine which is best suited with respect to the specific problem domain that is being addressed. 

Spatial-temporal resource allocation has a key property that informs the solution method choice --- a highly dynamic environment that is difficult to model in closed-form. To illustrate, consider an agent travel model. While there are certainly long term trends for travel times, precise predictions are difficult due to complex interactions between features such as traffic, weather, and events occurring in the city. A city's traffic distribution also changes over time as the road network and population shifts, so it needs to be updated periodically with new data. This dynamism is true for many pieces of the domain's environment, including the demand distribution of incidents. Importantly, it is also true of the system itself: agents can enter and leave the system due to mechanical issues or purchasing decisions, and depots can be closed or opened. 

Whenever underlying environmental models change, the solution approach must take the updates into account to make correct recommendations. Approaches that require long training periods such as reinforcement learning and value iteration are difficult to apply since they must be re-trained each time the environment changes. This motivates using Monte Carlo Tree Search (MCTS), a probabilistic search algorithm, as our solution approach. Being an anytime algorithm, MCTS can immediately incorporate any changes in the underlying generative environmental models when making decisions. 

% evaluate actions by sampling from a large number of possible scenarios.
MCTS represents the planning problem as a ``game tree'', where states are represented by nodes in the tree. The decision-maker is given a state of the world and is tasked with finding a promising action for the state. The current state is treated as the root node, and actions that take you from one state to another are represented as edges between corresponding nodes. The core idea behind MCTS is that this tree can be explored asymmetrically, with the search being biased toward actions that appear promising. To estimate the value of an action at a state node, MCTS simulates a ``playout'' from that node to the end of the planning horizon using a computationally cheap \textit{default policy} (our simulated system model is shown in figure \ref{fig:state} and described in detail in appendix \ref{sec:app_sim}). This policy is generally not very accurate (a common method is random action selection), but as the tree is explored and nodes are revisited, the estimates are re-evaluated and will converge toward the true value of the node. This asymmetric tree exploration allows MCTS to search very large action spaces quickly. 
% To evaluate potential actions, MCTS simulates the system model, which is shown in figure \ref{fig:state} and described in detail in appendix \ref{sec:app_sim}.Each system state consists of a queue of pending incidents and a set of agents, as described in section \ref{sec:problem_formulation}. Our planner has access to several actuation functions. For example, the $\text{Dispatch}(\lambda_j, incident)$ function assigns agent $\lambda_j$ to the pending incident by updating it's destination cell to the location of the incident, updating its status to \textit{<busy>}, and returning the response time to the incident. Travel times when moving to a new location are determined by using an exogenously provided traffic routing model. 

When implementing MCTS, there are a few domain specific decisions to make --- the \textit{tree policy} used to navigate the search tree and find promising nodes to expand, and the \textit{default policy} used to quickly simulate playouts and estimate the value of a node. 

\textbf{Tree Policy:} When navigating the search tree to determine which nodes to expand, we use the standard Upper Confidence bound for Trees (UCT) algorithm \cite{kocsis2006bandit}, which defines the score of a node $n$ as 
\begin{equation}
    \text{UCB}(n) = \overline{u}(n) + c\sqrt{\frac{\text{log}(\text{visits}(n))}{\text{visits}(n')}}
\end{equation}
where $\overline{u(n)}$ is the estimated utility of state at node $n$, visits($n$) is the number of times $n$ has been visited, and $n'$ is $n$'s parent node. When deciding which node to explore in the tree, the child node with the maximum UCB score is chosen. The left term $\overline{u(n)}$ is the exploitation term, and favors nodes that have been shown to be promising. The right term is the exploration term, and benefits nodes with low visit counts, which encourages the exploration of under-represented actions in the hope of finding an overlooked high value path. The constant $c$ controls the tradeoff between these two opposing objectives, and is domain dependent. 

\textbf{Default Policy:} When working outside the MCTS tree to estimate the value of an action, i.e. rolling out a state, a fast heuristic \textit{default policy} is used to estimate the score of a given action. Rather than using a random action selection policy, we exploit our prior knowledge that agents generally stay at their current depot unless large shifts in incident distributions occur. Therefore, we use greedy dispatch without any redistribution of responders as our heuristic default policy.

It is important to note that performing MCTS on one sampled chain of events is not enough, as traffic incidents are inherently sparse. Any particular sample will be too noisy to make robust claims regarding the value of an action. To handle this uncertainty, we use \textit{root parallelization}. We sample many incident chains from the prediction model and instantiate separate MCTS trees to process each. We then average the scores across trees for each potential allocation action to determine the optimal action. 

Our low-level planning approach is shown in algorithm \ref{algo:low}. The inputs for low-level planning are the regions $R$, the current overall system state $s$ (which includes each agent's region assignment), a generative demand model $E$, and the number of chains to sample and average over for each region $n$. For each region $r_j \in R$, we first extract the state $s_j$ in the region's MSMDP from the current overall system state $s$ (step ~\ref{algo:low_decompose}). Then we perform root parallelization by sampling $n$ incident chains from the demand model $E$ and performing MCTS on each to score each potential allocation action (step ~\ref{algo:low_sampleChains}). It is important to note that the sampled incident chains are specific to the region under investigation, and demand is only generated from the cells that are in that region. We then average the scores across samples for each action, and choose the allocation action with the maximum average score (step ~\ref{algo:low_argmax}).

\begin{algorithm}[t]
\SetAlgoLined
\SetKwInOut{Input}{input}
\SetKwInOut{Output}{output}
\Input{Regions $R$, State $s$, Generative Demand Model $E$, Number of Samples $n$}
\Output{Recommended Allocation Actions $\sigma_r \,\, \forall r \in R$}
 
 \For{$\text{region } r_j \in R$}{
    $\text{Decompose } s \text{ into region specific state } s_j$\;\label{algo:low_decompose}
    $\text{Action Score Map } \widetilde{\mathcal{A}} := \emptyset$\;
    % \While{$i < n$}{
    $\text{eventChains } := E\text{.sample}(s_j, n)$\;\label{algo:low_sampleChains}
    $\text{action scores } A := \text{MCTS}(s_j, \text{eventChains})$\;
    \For{$\text{action }a \in A$}{
        $\widetilde{\mathcal{A}}[a]\text{.append}(\text{score}(a))$\;
        % }
    }
    $\overline{\mathcal{A}} := \emptyset$\;
    \For{$\text{potential action }a \in A$}{
        $\overline{\mathcal{A}}[a] = \text{mean}(\widetilde{\mathcal{A}}[a])$\;
    }
    
    $\text{Recommended action }\sigma_r := \text{argmax}_{a} \ \ \overline{\mathcal{A}}[a]$\label{algo:low_argmax}

 }

 \caption{Low-Level Planner}
 \label{algo:low}
\end{algorithm}

% \begin{algorithm}
% \SetAlgoLined
% \SetKwInOut{Input}{input}
% \SetKwInOut{Output}{output}
% \Input{Regions $R$, State $s$, Generative Demand Model $E$, Number of Samples $n$}
% \Output{Recommended Allocation Actions $\sigma_r \forall r \in R$}
 
%  \For{$\text{region } r \in R$}{
%     $\text{Decompose } s \text{ into region specific state } s_r$\;
%     $\text{Action Score Map } \widetilde{\mathcal{A}} := \emptyset$\;
%     $i := 0$\;
%     \While{$i < n$}{
%         $\text{eventChain } := E\text{.sample}(s_r)$\;
%         $\text{action scores } a := \text{MCTS}(s_r, \text{eventChain})$\;
%     }
%  }

%  \caption{Low-Level Planner 2}
%  \label{algo:low}
% \end{algorithm}

%  $\text{assigned} := 0, i := 0, j := 0$\;
%  \While{\text{assigned} $\leq \,\,\mid \Lambda \mid \,\, \textbf{and} \,\, i \leq k$}{
%   $x_i := x_i + 1$\;
%   $\text{assigned} := \text{assigned} + 1$\;
%   \If{$w_i(x_i) < \infty$ }{
%   $i := i + 1$\;
%   }
%  }
%  \While{\text{assigned} $\leq \,\,\mid \Lambda \mid$}{
%   $\text{Calculate } G \text{ where } g_i = w_i(x_i + 1) - w_i(x_i)$\;
%   $r^{*} = \argmax_{r_i \in R}G$\;
%   $x_{r^{*}} := x_{r^{*}} + 1$\;

\section{Experiments} \label{sec:exp}

\begin{figure*}
    \centering
    \includegraphics[width=0.9\textwidth]{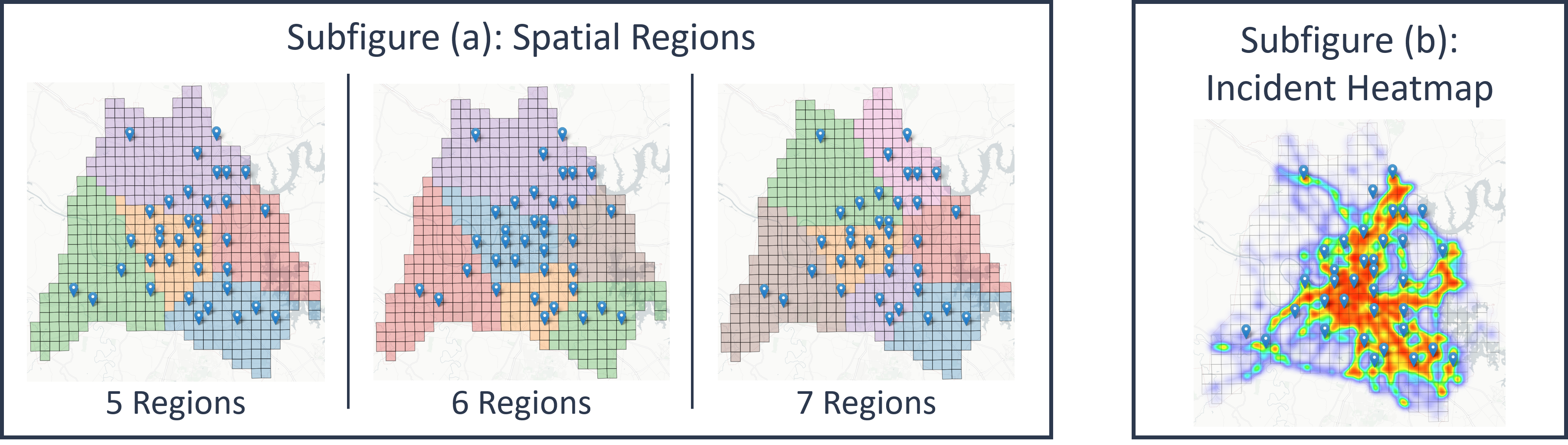}
    \caption{Subfigure (a) -- The various spatial regions under consideration. Pins on the map represent depot locations, and different colors represent different spatial regions. Subfigure (b) -- Nashville's historic incident density from January 2018 to May 2019 overlaid on the spatial grid environment.}
    \label{fig:combined_heatmap_regions}
\end{figure*}

We evaluate the proposed hierarchical framework's effectiveness on emergency response data obtained from Nashville, Tennessee, a major metropolitan area in the United States, with a population of approximately 700,000. We use historical incident data, depot locations, and operational data provided by the Nashville Fire Department~\cite{fireDepartmentCommunication}. We construct a grid representation of the city using $1\times1$ mile square cells. This choice was a consequence of the fact that a similar granularity of discretization is followed by local authorities. These cells, as well as the city's 35 depot locations, can be seen in figure~\ref{fig:combined_heatmap_regions}.

We make a few important assumptions when configuring our experiments. First, we limit the capacity of each depot $C(d)$ to 1. This encourages responders to be geographically spread out to respond quickly to incidents occurring in any region of the city, and it models the usage of ad-hoc stations by responders, which are often temporary parking spots.\footnote{In theory, we could always add dummy depots at the same location to extend our approach to a situation where more than one responder per depot is needed.} We assume there are 26 available responders to allocate, which is the actual number of responders in the urban area under consideration~\cite{fireDepartmentCommunication}. 
% Last, in our experiments we assume that responders travel ``as the crow flies'', i.e. travel times between cells are determined by euclidean distance. If deployed, this should be replaced by a travel model that incorporates the roadway network and traffic to find more realistic travel times. 

Our experimental hyper-parameter choices are shown in table ~
\ref{tab:hyperparams}. In our experiments, we vary the number of spatial regions to examine how their size and distribution effects performance of the hierarchical planner; the resulting region configurations can be seen in figure \ref{fig:combined_heatmap_regions}. We use the k-means algorithm~\cite{macqueen1967some} implemented in scikit-learn~\cite{scikit-learn} on historical incident data provided by the Tennessee Department of Transportation, which consists of $47862$ incidents that occurred from January 2018 to May 2019 in Nashville.

We assume that the mean rate to service an incident is 20 minutes based on actual service times in practice in Nashville (we hold this constant in our experiments to directly compare the planning approaches). As mentioned in section ~\ref{sec:approach}, we assume that incidents are homogeneous. The number of MCTS iterations performed when evaluating potential actions on a sampled incident chain is set to 1000 and the number of samples from the incident model that are averaged together using root parallelization during each decision step is set to 50. We run the hierarchical planner after each test incident to re-allocate responders. Further, if the planner is not called after a pre-configured time interval, we call it to ensure that the allocations are not stale. In our experiments this maximum time between allocations is set to 60 minutes. We ran experiments on an Intel i9-9980XE, which has 38 logical processors running at a base clock of 3.00 GHz, and 64 GB RAM.

\definecolor{Gray}{gray}{0.9}
\begin{table}[t]
\caption{Experimental hyper-parameter choices\label{tab:hyperparams}}
\centering
\small
\begin{tabular}{|l|l|}
\hline
\rowcolor{Gray}
\textbf{Parameter}                                                              & \textbf{Value(s)} \\ \hline
Number of Regions                                                               & \{5, 6, 7\}    \\ \hline
Maximum Time Between Re-Allocations                                             & 60 Minutes     \\ \hline
Incident Service Time                                                           & 20 Minutes     \\ \hline
Responder Speed                                                                 & 30 Mph         \\ \hline
MCTS Iteration Limit                                                            & 1000           \\ \hline
Discount Factor                                                                 & 0.99995        \\ \hline
UCT Tradeoff Parameter $c$                                                          & 1.44           \\ \hline
Number of Generated Incident Samples & 50             \\ \hline
% \begin{tabular}[c]{@{}l@{}}Number of Generated \\ Incident Samples\end{tabular} & 50             \\ \hline
\end{tabular}
%\vspace{-0.05in}
\end{table}

Our incident model is learned from the 47862 real incidents discussed earlier. For each cell, we learn a Poisson distribution over incident arrival based on  historical data. The maximum likelihood estimate (MLE) of the rate of the Poisson distribution is simply the empirical mean of the number of incidents in each unit of time. To simulate our system model, we access the Poisson distribution of each cell and sample incidents from it. In reality, emergency incidents might not be independently and identically distributed; however, the incident arrival model (and the blackbox simulator of the system in general) is completely exogenous to our model and does not affect the functioning of our approach. To validate the robustness of our approach, we create three separate test beds based on domain knowledge and preliminary data analysis of historical incident data.

% \begin{enumerate}[noitemsep,leftmargin=*]
\textbf{Stationary incident rates:} We start with a scenario where our forecasting model samples incidents from a Poisson distribution that is stationary (for each cell), meaning that the rate of incident occurrence for each cell is the empirical mean of historical incident occurrence per unit time in the cell. This means that the only utility of the high-level planner in such a case is to divide the overall spatial area into regions and optimize the initial distribution of responders among them. Since the rates are stationary, the initial allocation is maintained throughout the test period under consideration. This scenario lets us test the proposed low-level planning approach in isolation. The experiments were performed on five chains of incidents sampled from the stationary distributions, which have incident counts of \{$939, 937, 974, 1003, 955$\} respectively (for a total of 4808 incidents), and are combined to reduce noise.

\begin{figure}[t]
    \centering
    \includegraphics[width=0.8\columnwidth]{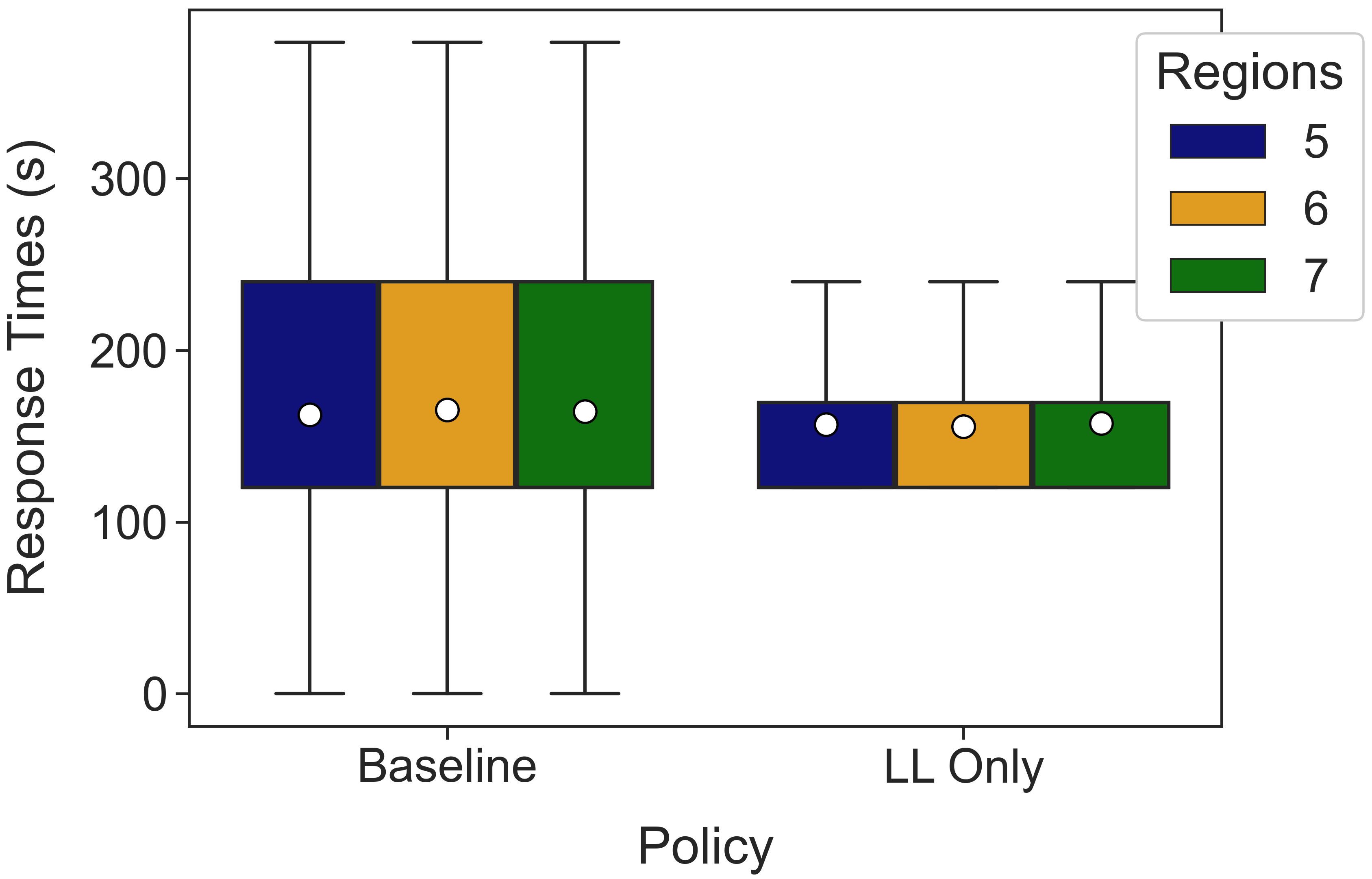}
    \caption{The response time distributions for the baseline and low-level planners (LL Only) when applied to incidents sampled from a stationary rate distribution. The boxplot represents the data's Inter-Quartile Range (IQR $ = Q_3 - Q_1$), and the whiskers extend to 1.5IQR.}
    \label{fig:results_stationary_dist}
% \end{figure}
% \begin{figure}
    \centering
    \includegraphics[width=0.8\columnwidth]{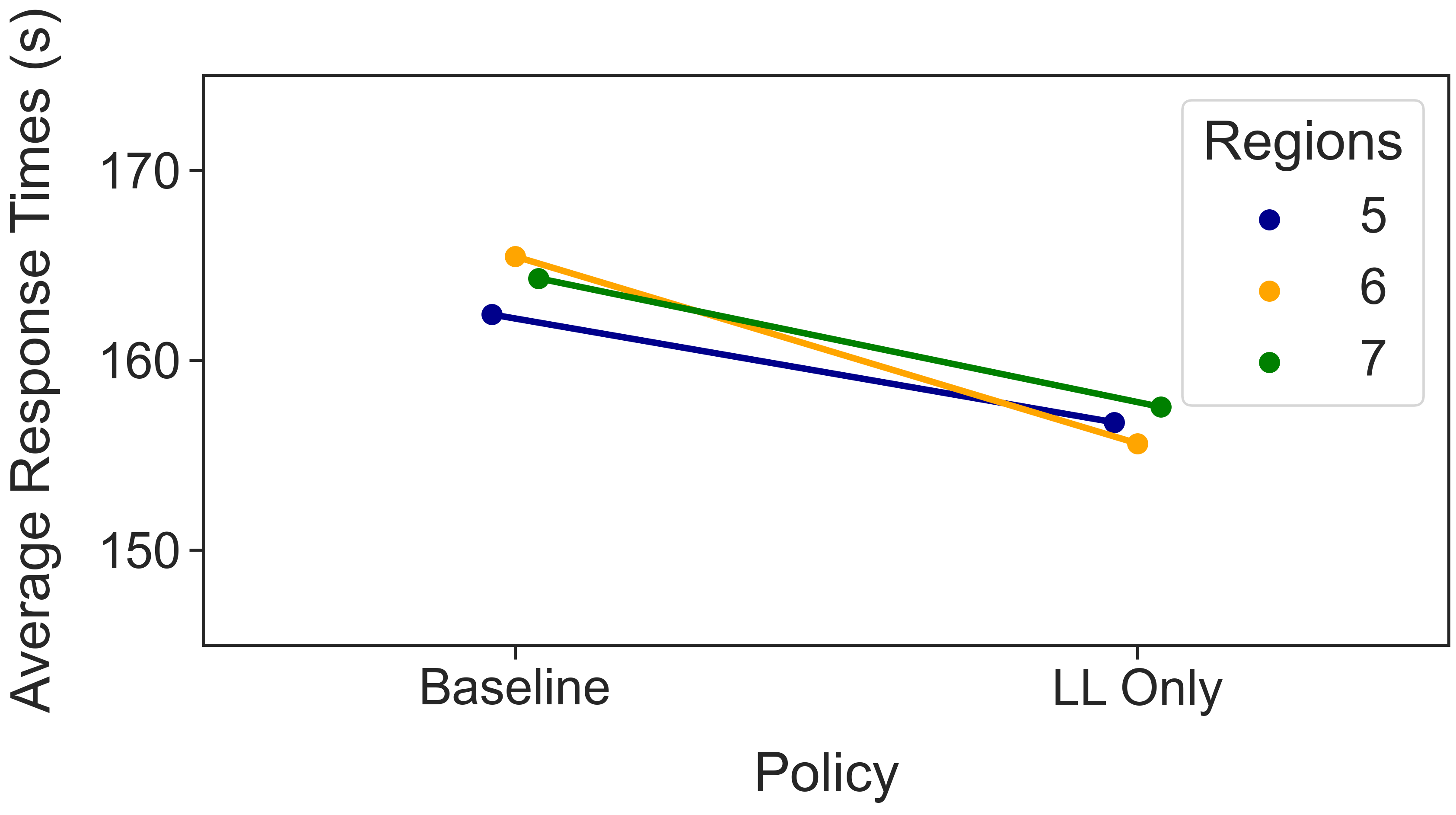}
    \caption{A zoomed in view of the average response times for the baseline and low-level planners (LL Only) when applied to incidents sampled from a stationary rate distribution.}
    \label{fig:results_stationary_avg}
    \vspace{-0.05in}
\end{figure}

\textbf{Non-stationary incident rates:} We test how our model reacts to changes in incident rates. We identify different types of scenarios that cause the dynamics of spatial temporal incident occurrence and traffic to change in specific areas of Nashville. We look at rush-hour traffic on weekdays (which affects the center of the county), football game days (which affects the area around the football stadium, typically on Saturdays), and Friday evenings (which affects the downtown area). Then, we synthetically simulate spikes in incident rates in the specific areas at times when the areas are expected to see spikes. To further test whether our approach can deal with sudden spikes, we randomly sample the spikes from a Poisson distribution with a rate that varies between two to five times the historical rates of the regions. We create five different trajectories of incidents with varying incident rates, which have incident counts of \{$873, 932, 865, 862, 883$\} respectively (for a total of 4415 incidents). In these experiments we compare using the low-level planner with fixed responder distributions across regions to a full deployment that incorporates the high-level planner to dynamically balance responders across regions. 

\textbf{Responder failures:} An important consideration in emergency response is to quickly account for situations where some ambulances might be unavailable due to maintenance and breakdowns. We randomly simulate failures of ambulances lasting 8 hours to understand how our approach deals with such scenarios.
% \end{enumerate}

We compare our approach with a baseline policy that has no responder re-allocation. This baseline emulates current policies in use by cities in which responders are statically assigned to depots and rarely move. The initial responder placement is determined using our proposed high-level policy to ensure all the policies begin with similar responder distributions. The baseline uses the same greedy dispatch policy as our approach.

\section{Results} \label{sec:exp_results}

\textbf{Stationary Incident Rates:} The results of experiments comparing the baseline policy with the proposed low-level planner on incidents sampled from stationary incident rates are shown in figures \ref{fig:results_stationary_dist} and \ref{fig:results_stationary_avg}. Our first observation is that using the low-level planner reduces response times for all region configurations, improving upon the baseline by \textbf{7.5} seconds on average. This is a significant improvement in the context of emergency response since it is well-known that paramedic response time affects short-term patient survival~\cite{mayer1979emergency}. We also observe a significant shift in the distribution of response times, with the upper quartile of the low-level results being reduced by approximately \textbf{71 seconds} for each region configuration. This reduction in variance indicates that the proposed approach is more consistent. As a result, lesser number of incidents experience large response times. 
% Our last observation is that there is little difference between the region configurations when using either the baseline or low level planner. These results demonstrate the effectiveness of our low level planning approach in isolation. 

\begin{figure}[t]
    \centering
    \includegraphics[width=0.8\columnwidth]{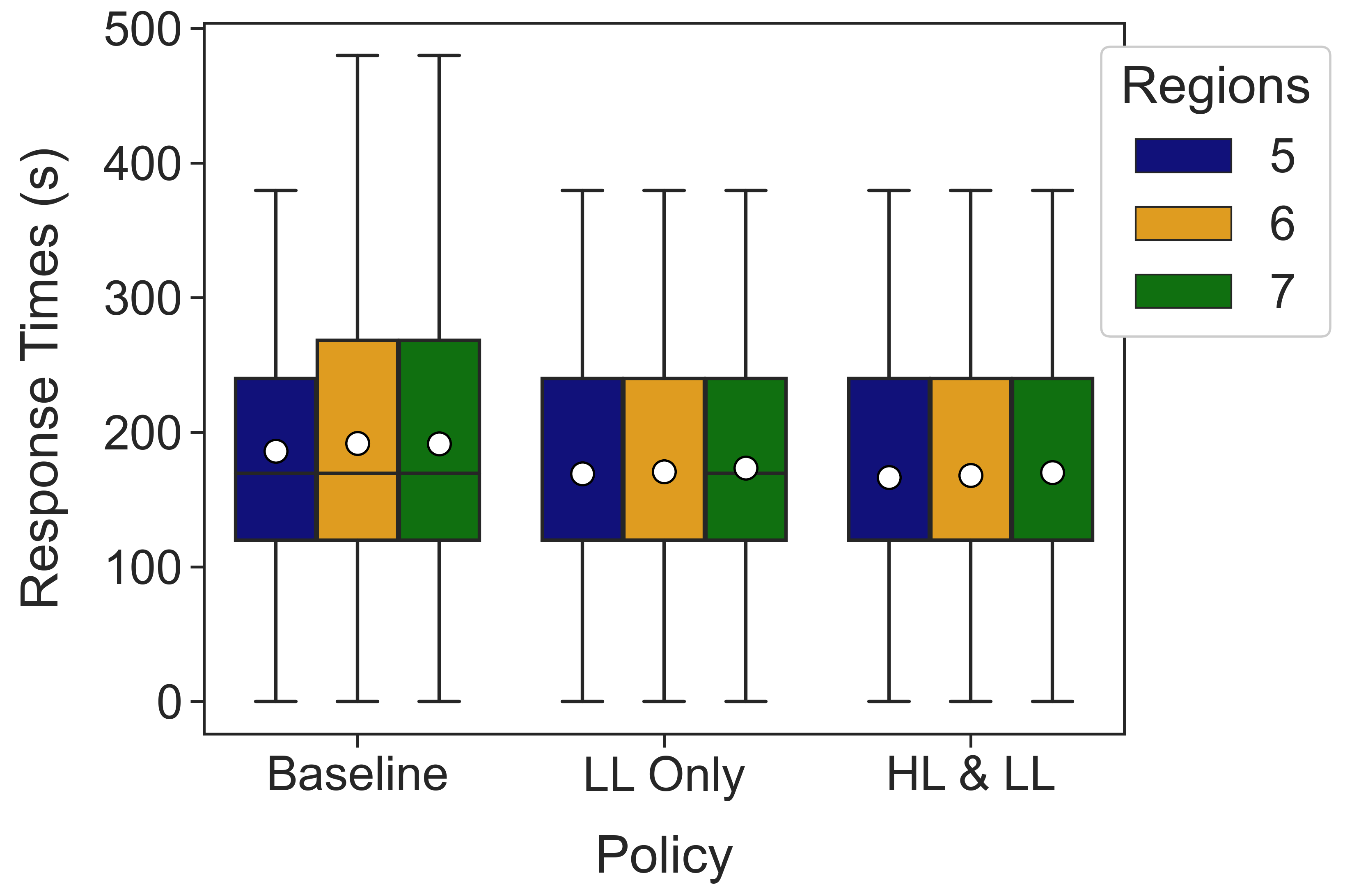}
    \caption{The response time distributions for the baseline, low-level planner (LL Only), and complete hierarchical planner (HL \& LL) when applied to incidents sampled from a non-stationary rate distribution.}
    \label{fig:results_non_stationary_dist}
% \end{figure}
% \begin{figure}
    \centering
    \includegraphics[width=0.8\columnwidth]{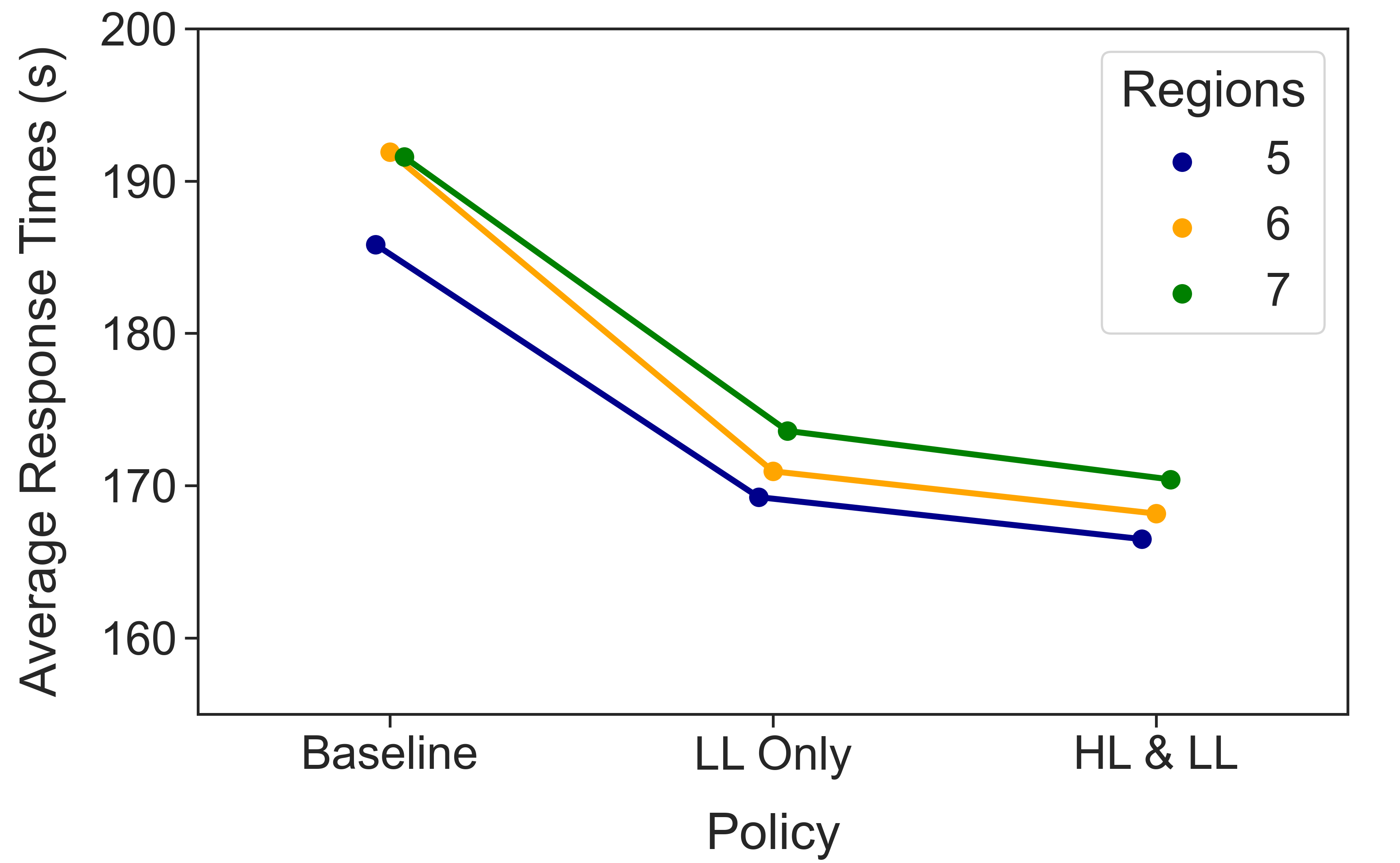}
    \caption{A zoomed in view of the average response times for the baseline, low-level planner (LL Only), and complete hierarchical planner (HL \& LL) when applied to incidents sampled from a non-stationary rate distribution.}
    \label{fig:results_non_stationary_avg}
  %  \vspace{-0.05in}
\end{figure}

\textbf{Non-Stationary Incident Rates:} We now examine the results of experiments using incidents generated from non-stationary incident distributions, which are shown in figures \ref{fig:results_non_stationary_dist} and \ref{fig:results_non_stationary_avg}. Our first observation is that response times generally increase relative to the stationary experiments for both the baseline and the proposed approach. This result is expected since response to incidents sampled from a non-stationary distribution are more difficult to plan for. However, we also observe that our approach is better able to adapt to the varying rates. The low-level planner in isolation improves upon the baseline's response times by \textbf{18.6 seconds} on average. Introducing the complete hierarchical planner (i.e. both the high-level and low-level planners) improves the result further, reducing response times by \textbf{3 seconds} compared to using only the low-level planner, and \textbf{21.6 seconds} compared to the baseline. We again observe that the region configuration has a small effect on the efficiency of the proposed approach. This result shows that our approach reduces lower response times irrespective of the manner in which the original problem is divided into regions. Finally, we also observe that the variance of the response time distributions achieved by the proposed method is not as low as compared to the stationary experiments, which is likely due to the high strain placed on the system from the non-stationary incident rates.

% 189.817938  164.06
% 168.291791  156.6
% 171.209525

% \begin{figure}
%      \centering
%      \begin{subfigure}[c]{0.95\columnwidth}
%          \centering
%          \includegraphics[width=\textwidth]{figures/failure_resp_time_results.png}
%          \caption{Response time distributions}
%          \label{fig:results_failure_dist}
%      \end{subfigure}
%      \hfill
%      \begin{subfigure}[c]{0.95\columnwidth}
%          \centering
%          \includegraphics[width=\textwidth]{figures/failure_avg_resp_time_results.png}
%          \caption{Zoomed in comparison of average response times}
%          \label{fig:results_failureavg}
%      \end{subfigure}
%         \caption{Non-stationary results}
%         \label{fig:results_failure}
% \end{figure}

\begin{figure}[t]
    \centering
    \includegraphics[width=0.75\columnwidth]{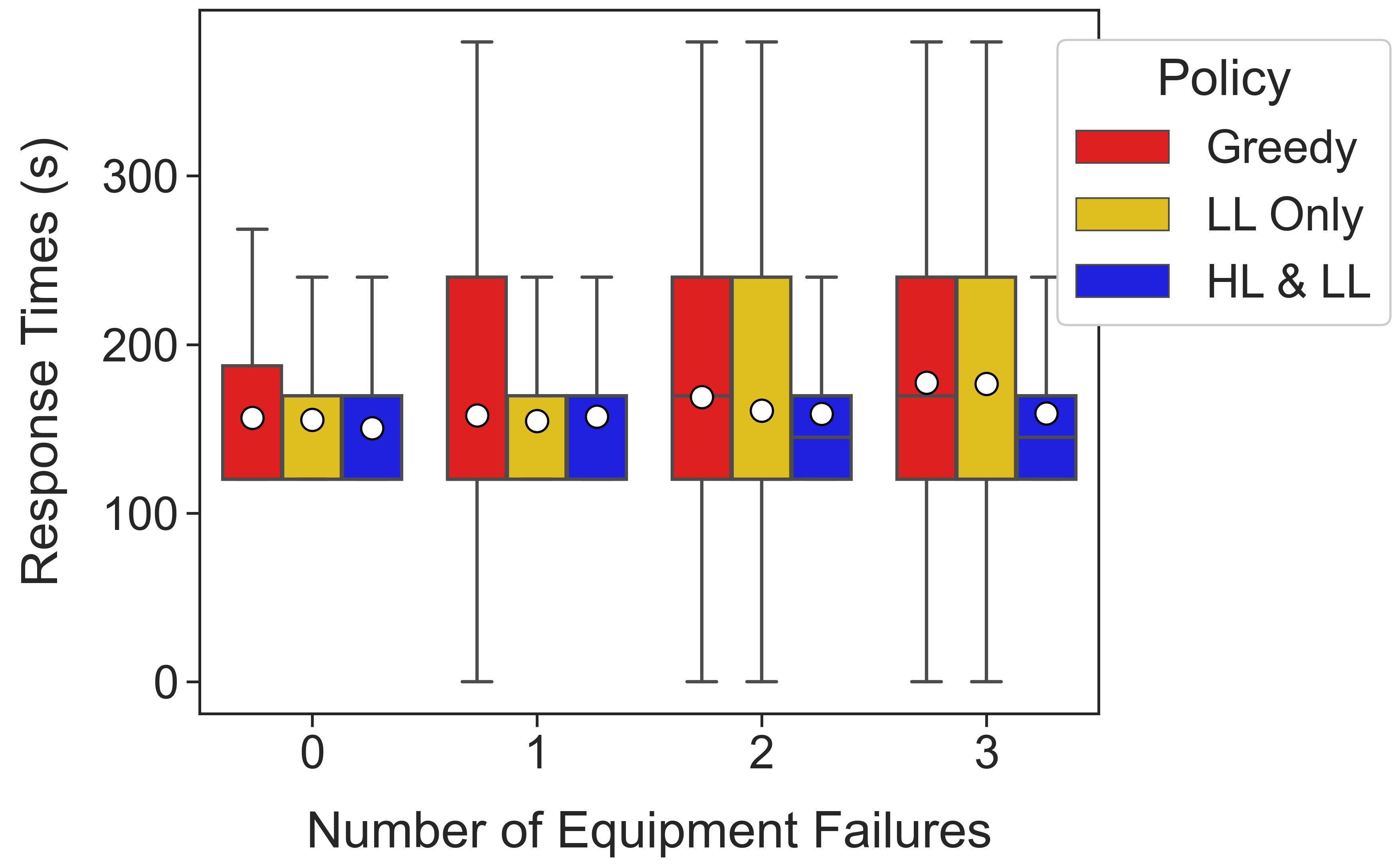}
    \caption{The response time distributions for the baseline, low-level planner (LL Only), and complete hierarchical planner (HL \& LL) when subjected to increasing numbers of simultaneous equipment failures.}
    \label{fig:results_failure_dist}
% \end{figure}
% \begin{figure}
    \centering
    \includegraphics[width=0.75\columnwidth]{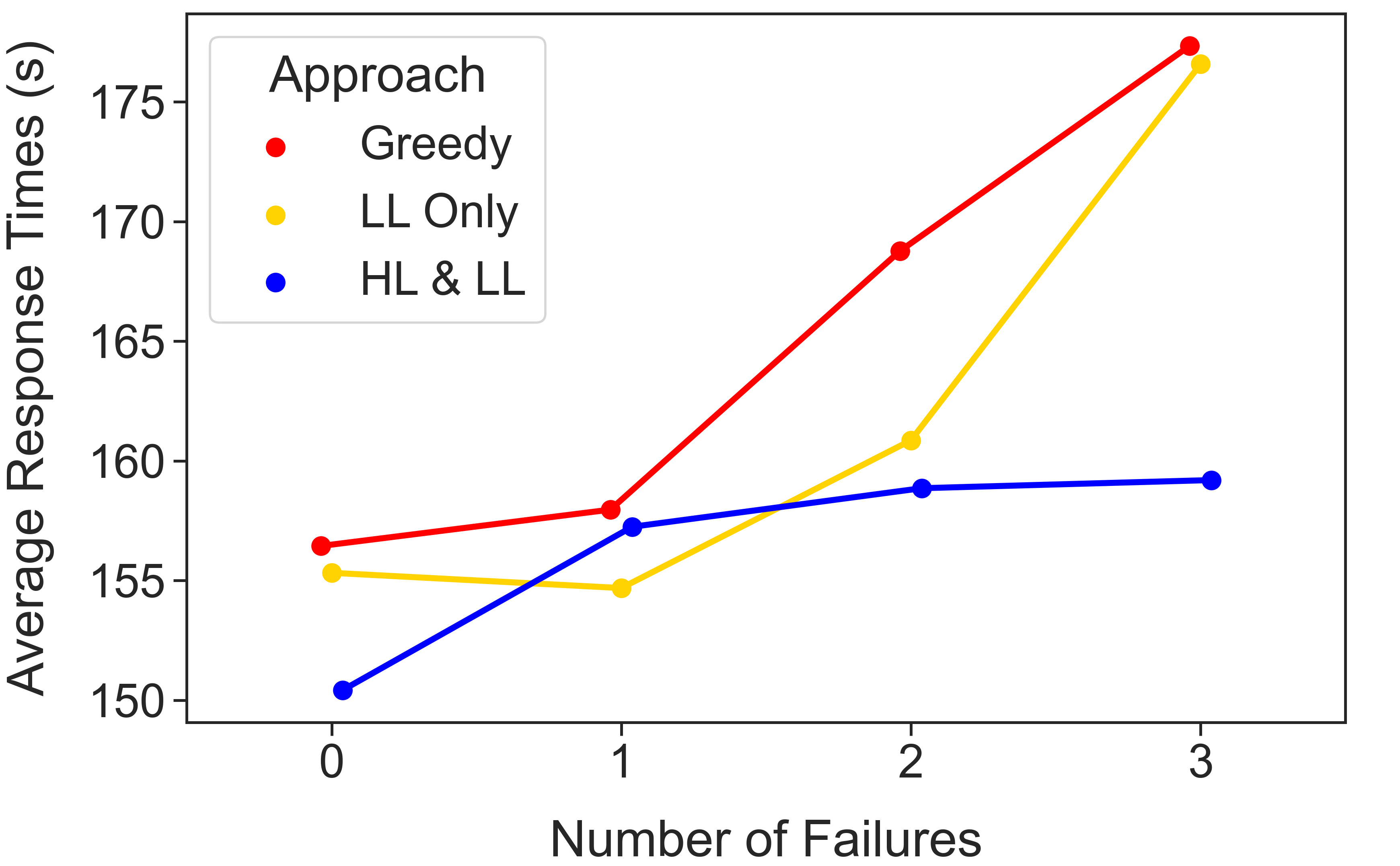}
    \caption{A zoomed in view of the average response times for the baseline, low-level planner (LL Only), and complete hierarchical planner (HL \& LL) when subjected to increasing numbers of simultaneous equipment failures.}
    \label{fig:results_failureavg}
  
 %   \vspace{-0.1in}
\end{figure}

\begin{figure}[t]
    \centering
    \includegraphics[width=.88\columnwidth]{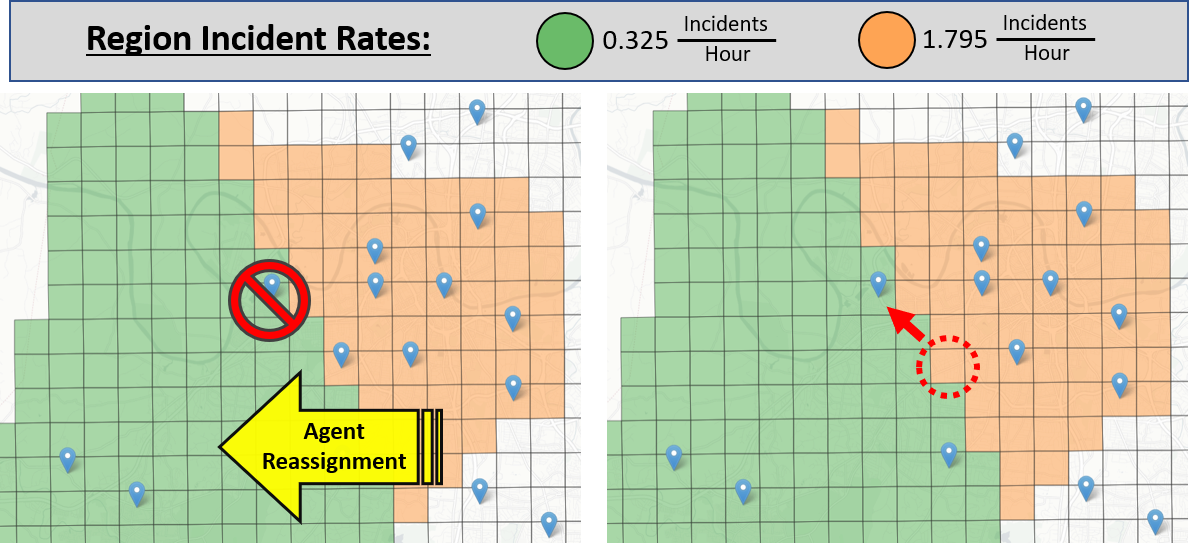}
    \caption{Example of the high-level planner resolving an equipment failure. In sub-figure (left), the agent positioned at the depot marked by the red circle in the green region fails, and the high-level planner determines there is an imbalance across regions. In sub-figure (right), we see the planner move an agent from the depot marked by the red dotted circle to the green region to ensure that the upper left of the region can be serviced. }
    \label{fig:failure}
% \end{figure}

   \centering
    \includegraphics[width=0.75\columnwidth]{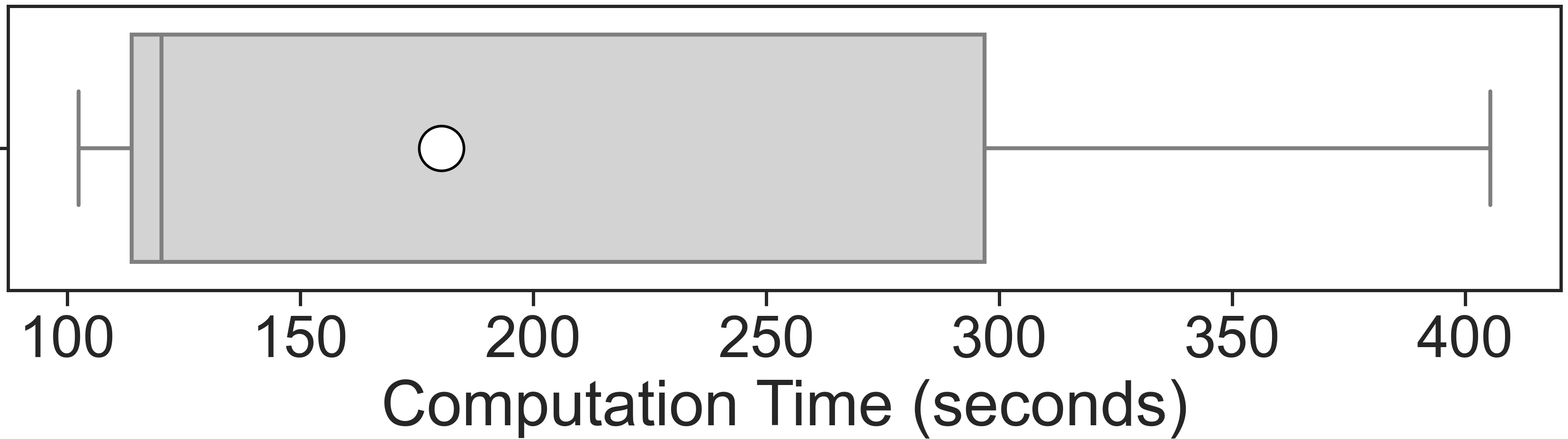}
    \caption{Distribution of Computation times}
    \label{fig:compu_time}

%  \begin{figure}[t]
   
%    \vspace{-0.2in}
\end{figure}

\textbf{Responder Failures: } Results on the non-stationary incident distribution demonstrate the effectiveness of the hierarchical planner when there are shifts in the spatial distribution of incidents. We now examine its response to equipment failures within the ERM system. Figure \ref{fig:failure} illustrates an example (from our experiments) of how the planner can adapt to equipment failures. When a responder in the green region fails, the high-level planner determines that imbalance in the spatial distribution of the responders. Intuitively, due to the failure incidents occurring in the upper left cells of the green region could face long response times. Therefore, the planner reallocates a responder from the orange region to the green region. 

To examine how equipment failures impact the proposed approach, we simulated several responder failures and compared system performance using our three allocation strategies. We show the results in figures \ref{fig:results_failure_dist} and \ref{fig:results_failureavg}. Naturally, as the number of failures increases, response times increase as there are fewer responders. However, we observe that the proposed approach intelligently allocates the remaining responders to outperform the baseline methods. Indeed, when there are three simultaneous failures, using the hierarchical planning improves response times by over \textbf{20 seconds} compared to both the baseline and using only the low-level planner.

% \begin{figure}
%      \centering
%      \begin{subfigure}[t]{0.49\columnwidth}
%          \centering
%          \includegraphics[width=\textwidth]{figures/failure_demo_1_label.png}
%          \caption{Failure Occurrence}
%          \label{fig:failure_1}
%      \end{subfigure}
%      \hfill
%      \begin{subfigure}[t]{0.49\columnwidth}
%          \centering
%          \includegraphics[width=\textwidth]{figures/failure_demo_2.png}
%          \caption{Failure Resolution}
%          \label{fig:failure_2}
%      \end{subfigure}
%         \caption{Example of the high level planner resolving an equipment failure. In sub-figure (a), the agent positioned at the depot marked by the red circle in the green region fails, and the High level planner determines there is an imbalance across regions. In sub-figure (b), we see the planner move an agent from the depot marked by the red dotted circle to the green region to ensure that the upper left of the region can be serviced. }
%         \label{fig:failure}
% \end{figure}

% Example of the high level planner resolving an equipment failure. In sub-figure (a), the agent positioned at the depot marked by the red circle in the green region fails. Despite the green region having a lower incident rate than the orange region, the large area it services can lead to high wait times. The High level planner determines the failure has led to an imbalance and that an agent should be moved from the orange region to the green region. In sub-figure (b), we see that the planner chooses to move an agent from the depot marked by the red dotted circle to the depot that experienced the failure to ensure that the upper left of the region can be serviced in the event of an incident. 

\textbf{Allocation Computation Times: } %To be viable,
%The system
%For our proposed system to be viable for use by real world ERM systems, it 
%must be able to perform allocation computations in a reasonable amount of time.
The distribution of computation times can be seen in figure \ref{fig:compu_time}. Decisions using the proposed approach take  \textbf{180.29} seconds on average. Note that this is the time that our system takes to optimize the allocation of responders. Dispatch decisions are greedy and occur instantaneously. Hence, our system can easily be used by first responders on the field without hampering existing operational speed.

% To test this, we deployed the system on a high performance computer (an intel i9-9980XE based system, which has 38 logical processors running at a base clock of 3.00 GHz, and 64 GB RAM) and recorded the computation times for one day's worth of incidents (167). The distribution of computation times can be seen in figure \ref{fig:compu_time}. Each allocation took the system \textbf{180.29} seconds on average. 

\section{Related work}\label{sec:related}

Markov decision processes can be directly solved using dynamic programming when the transition dynamics of the system are known ~\cite{kochenderfer2015decision}. Typically, for resource allocation problems in complex environments like urban areas, the transition dynamics are unknown~\cite{mukhopadhyay2020review}. To alleviate this, our Simulate-and-Transform (\textit{SimTrans}) algorithm \cite{mukhopadhyayAAMAS18} can be used which performs canonical Policy Iteration with an added computation. In order to estimate values (utilities) of states, the algorithm simulates the entire system of incident occurrence and responder dispatch and keeps track of all states transitions and actions, and gradually builds statistically confident estimates of the transition probabilities.

While it finds a close approximation of the optimal policy (assuming that the estimates of the transition probabilities are close to the true probabilities), this process is extremely slow and unsuited to dynamic environments. As an example, even if a single agent (ambulance in this case) breaks down, the entire process of estimating transition probabilities and learning a policy must be repeated. To better react to dynamic environmental conditions, decentralized and online approaches have been explored~\cite{claes2017decentralised,MukhopadhyayICCPS}. For example, ~\citeauthor{claes2017decentralised}~\cite{claes2017decentralised} entrust each agent to build its own decision tree and show how computationally cheap models can be used by agents to estimate the actions of other agents as the trees are built.

An orthogonal approach to solve large-scale MDPs is using hierarchical planning~\cite{hauskrecht2013hierarchical}. Such an approach focuses on learning local policies, known as \textit{macros}, over subsets of the state space. The concept of macro-actions was actually introduced separately from hierarchical planning, as means to reuse a learned mapping from states to actions to solve multiple MDPs when objectives change~\cite{sutton1995td, precup1998multi}. Later, the macro-policies were used in hierarchical models to address the issue of large state and action spaces~\cite{forestier1978multilayer,hauskrecht2013hierarchical}. 

We also describe how allocation and dispatch are handled in the context of emergency response. First, note that the distinction between allocation and response problems can be hazy since any solution to the allocation problem implicitly creates a policy for response (greedy response based on the allocation)~\cite{mukhopadhyay2020review}. We use a similar approach in this paper since greedy response satisfies the constraints under which first responders operate. The most commonly used metric for optimizing resource allocation is coverage~\cite{toregas_location_1971,church1974maximal, gendreau_solving_1997}. Waiting time constraints are often used as constraints in approaches that maximize coverage~\cite{silva2008locating,mukhopadhyayAAMAS17}. Decision-theoretic models have also been widely used to design ERM systems. For example, \citeauthor{keneally2016markov}~\cite{keneally2016markov} model the resource allocation and dispatch problem in ERM as a continuous-time MDP, while we have previously used a semi-Markovian process~\cite{mukhopadhyayAAMAS18}. Allocation in ERM can also be addressed by optimizing distance between facilities and demand locations~\cite{MukhopadhyayICCPS}, and explicitly optimizing for patient survival~\cite{erkut_ambulance_2008, knight_ambulance_2012}.
%\vspace{-0.2in}
\section{Conclusion} \label{sec:conclusion}

We have presented a hierarchical planning approach for dynamic resource allocation in city scale cyber-physical system (CPS). We model the overall problem as a Multi-Agent Semi-Markov Decision Process (MSMDP), and show how to leverage the problem's spatial structure to decompose the MSMDP into smaller and tractable sub-problems. We then detail how a hierarchical planner can employ a low-level planner to solve these sub-problems, while a high-level planner identifies situations in which resources must be moved across region lines.  Our experiments show that our proposed hierarchical approach offers significant improvements when compared to the state-of-the-art in emergency response planning, as it maintains system fairness while significantly decreasing average incident response times. We also find that it is robust to equipment failure and is computationally efficient enough to be deployed in the field without hampering existing operational speed. While this work demonstrates the potential of hierarchical decision making, several non-trivial technical challenges remain, including how to optimally divide a spatial area such that the solutions of the sub-problems maximize the overall utility of the original problem. We will explore these challenges in future work.
\bibliographystyle{ACM-Reference-Format}
\bibliography{references}
\balance
%------------------------------------------------
% Appendix
%------------------------------------------------

\newpage
% \newpage
\appendix
\appendixpage

\section{Simulator}
\label{sec:app_sim}

This appendix describes our simulation of the emergency response system and the environment used by the low-level planner to estimate the effect of various actions. As shown in figure \ref{fig:state}, our system state at time $t$ is captured by a queue of active incidents $I^{t}$ and agent states $\Lambda$. $I^t$ is the queue of incidents that have been reported but not yet serviced, and allows the system to keep track of any incidents that could not be immediately responded to. The state of each agent $\lambda_j \in \Lambda$ consists of the agent's current location $p_j^t$, status $u_j^t$ destination $g_j^t$, assigned region $r_j^t$, and assigned depot $d_j^t$. Each agent can be in several different internal states (represented by $u_j^t$), including \textit{waiting} (waiting at a depot), \textit{in\_transit} (moving to a new depot and not in emergency response mode), \textit{responding} (the agent has been dispatched to an incident and is moving to its location), and \textit{servicing} (the agent is currently servicing an incident). These states dictate how the agent is updated when moving the simulator forward in time. 

To determine the travel times between locations in the environment, the simulator uses a traffic router. In our experiments, we use a Euclidean distance based router, which assumes all agents travel in straight lines between locations. If deployed to a real world system, a more advanced router can be used that uses information about the roadway network and current traffic conditions to accurately estimate travel times. 

Our simulator is designed as a discrete event simulator, meaning that the state is only updated at discrete time steps when interesting events occur. These events include incident occurrence, re-allocation planning steps, and responders becoming available for dispatch. Between these events, the system evolves based on prescribed rules. Using a discrete event simulator saves on valuable computation time as compared to a continuous time simulator. 

At each time step when the simulator is called, the system's state is updated to the current time of interest. First, if the current event of interest is an incident occurrence, it is added to the active incidents queue $I^t$. Then each agent's state and locations are updated to where they would be at the given time, which depends on their current state. For example, agents that are in the \textit{waiting} state stay at the same position, while agents that are \textit{responding} or \textit{in\_transit} will check to see if they have reached their destination. If they have, they will update their state to \textit{servicing} or \textit{waiting} respectively and update their locations. If they have not reached their destination, they interpolate their current location using the travel model. If an agent is in the \textit{servicing} state and finishes servicing an incident, it will enter the \textit{in\_transit} state and set its destination $g_j^t$ to its assigned depot. 

After the state is updated, a planner has several actuation's available to control the system. The \textit{Dispatch}$(\lambda_j, \textit{incident})$ function will dispatch the agent $\lambda_j$ to the given incident which is in $I^t$. Assuming the responder is available, the system sets $\lambda_j$'s destination $g_j^t$ to the incident's location, and its status $u_j^t$ is set to \textit{responding}. The incident is also removed from $I^t$ since it is being serviced, and the response time is returned to the planner for evaluation. The planner can also change the allocation of the agents.

\textit{AssignRegion}$(\lambda_j, r_j)$ assigns agent $\lambda_j$ to region $r_j$ by updating $\lambda_j$'s $r_j^t$. \textit{AssignDepot}$(\lambda_j, d_j)$ similarly assigns agent $\lambda_j$ to depot $d_j$ by updating $\lambda_j$'s $d_j^t$ and setting its destination $g_j^t$ to the depots location. These functions allow a planner to try different allocations and simulate various dispatching decisions. 

\section{Notation}

We summarize notation in table \ref{tab:lookup-table}.  

% Please add the following required packages to your document preamble:
% \usepackage{graphicx}
% \usepackage[normalem]{ulem}
% \useunder{\uline}{\ul}{}

\definecolor{Gray}{gray}{0.9}
\begin{table}[h!]
\caption{Notation lookup table}
\vspace{-0.1in}
\footnotesize
%  \fontsize{5}{6}\selectfont
% \tiny
\captionsetup{font=small}
% \small
% \fontsize{5.0pt}{5.0pt}\selectfont
\resizebox{.9\columnwidth}{!}{%
\begin{tabular}{|l|l|}
\hline
\rowcolor{Gray}
{\ul \textbf{Symbol}}     & {\ul \textbf{Definition}}                                                                       \\ \hline
$\Lambda$                 & Set of agents                                                                                   \\ \hline\rowcolor{Gray}
$D$                       & Set of depots                                                                                   \\ \hline
$\mathcal{C}(d)$          & Capacity of depot $d$                                                                           \\ \hline\rowcolor{Gray}
$G$                       & Set of cells                                                                                    \\ \hline
$R$                     & Set of regions                                                                                    \\ \hline\rowcolor{Gray}
$S$                       & State space                                                                                     \\ \hline
$A$                       & Action space                                                                                    \\ \hline\rowcolor{Gray}
$P$                       & State transition function                                                                       \\ \hline
$T$                       & Temporal transition distribution                                                                \\ \hline\rowcolor{Gray}
$\alpha$                  & Discount factor                                                                                 \\ \hline
$\rho(s, a)$                    & Reward function given action $a$ taken in state $s$                                       \\ \hline\rowcolor{Gray}
$\mathcal{A}$             & Joint agent action space                                                                        \\ \hline
$\mathcal{T}$             & Termination scheme                                                                              \\ \hline\rowcolor{Gray}
$s^t$                     & Particular state at time $t$                                                                    \\ \hline
$I^t$                     & Set of cell indices waiting to be serviced                                                     \\ \hline\rowcolor{Gray}
$\mathcal{Q}(\Lambda)$    & Set of agent state information                                                               \\ \hline
$p^t_j$                   & Position of agent $j$                                                                           \\ \hline\rowcolor{Gray}
$g^t_j$ & Destination of agent $j$                                                                                            \\ \hline
$u^t_j$ & Current status of agent $j$                                                                                      \\ \hline\rowcolor{Gray}
$s_i, s_j$                & Individual states                                                                               \\ \hline
% $t_{ij}$                    & Transition time between states $s_i, s_j$                                                       \\ \hline
% $t_a$                     & Time between incidents                                                                          \\ \hline\rowcolor{Gray}
% $t_s$                     & Time to service an incident                                                                     \\ \hline
% $t_r$                     & \begin{tabular}[c]{@{}l@{}}Incident response time (the time between incident \\ awareness and first agent's arrival on scene)\end{tabular} \\ \hline\rowcolor{Gray}
$\sigma$                  & Action recommendation set                                                                        \\ \hline\rowcolor{Gray}
$\eta$                    & Service Rate                                                                                     \\ \hline
$\gamma_g$                & Incident rate at cell $g$                                                                       \\ \hline\rowcolor{Gray}
$t_h$                     & Time since beginning of planning horizon                                                        \\ \hline
$t_r(s,a)$                   & Response time to an incident given action $a$ in state $s$                                      \\ \hline\rowcolor{Gray}

                                                            %   \\ \hline
\end{tabular}%
}
\vspace{-0.2in}
\label{tab:lookup-table}
\end{table}

\section{Reproducibility}

We have made the code used in this study publicly available.\footnote{\href{https://github.com/StatResp/Hierarchical_ERM_ICCPS}{https://github.com/StatResp/Hierarchical\_ERM\_ICCPS}} Detailed instructions on how to use the code can be found in the repository's readme. Below is an overview of the code and where to find various components.

\begin{itemize}
    \item \textbf{Decision\_making}: This folder contains the bulk of the decision process's implementation. \textit{MMCHighLevelPolicy.py} implements the high level planner. The \textit{./LowLevel/} folder contains the low level policy implementation, including the MCTS implementation, the reward function, and more. The \textit{./coordinator/} folder contains the various decision coordinator implementations, which control what decision strategy is used. For example, \textit{DispatchOnlyCoord.py} implements a static allocation policy. 
    
    \item \textbf{Environment}: This folder contains the simulator and information about the problem's environment. For example, \textit{ResponderDynamics.py} defines responders' behavior,  \textit{CellTravelModel.py} defines how long it takes to travel between locations, and \textit{Spatial/spatialStructure.py} defines the spatial grid which discretizes the environment.
    
    \item \textbf{Prediction}: This folder contains the incident forecasting model. 
    
    \item \textbf{Scenarios}: This folder contains definitions of different experimental scenarios that can be run. This includes defining the experimental environment, setting up the decision framework, and running the experiment. For example, \textit{./gridworld\_example/} defines a simple gridworld environment for demonstration purposes. 
    
\end{itemize}

To run an experiment, go to the appropriate scenario folder and run the experiment's script. This will define the environment, load the appropriate forecasting model, construct the desired decision framework, and then run a simulated ERM system using the desired allocation strategy. More information for how the framework is configured and how to define your own experiment can be found in the repository's readme. 

The incident data used in this study is proprietary, but we have released a synthesized example dataset (in the \textit{data/} folder) to demonstrate the expected data format. The incident data consists of a time series of incident events, each of which has a location and time of occurrence. We have provided a chain of synthetic `real incidents', which are used to evaluate the model, as well as several chains of incidents sampled from the forecasting model which are used in planning. We have run example experiments on this synthetic data, which are defined in the \textit{scenarios/synthesized/} folder, and have provided our results for reference.

\end{document}